\documentclass[journal]{IEEEtran}

\usepackage{graphicx}
\usepackage{multirow}
\usepackage{booktabs}
\usepackage{makecell}
\usepackage{amssymb}
\usepackage{mathrsfs}
\usepackage{amsmath}
\usepackage{cite}
\usepackage{url}
\usepackage{bbm}
\usepackage{bbding}
\usepackage[table]{xcolor}
\usepackage{xcolor}
\definecolor{rblue}{rgb}{0,0.5,1}
\usepackage{hyperref}
\hypersetup{colorlinks=true,linkcolor=red,citecolor=rblue}

\newcommand{\revised}[1]{{#1}}
\newcommand{\revisedT}[1]{{#1}}
\newcommand{\revisedF}[1]{\textcolor{black}{#1}}
\begin{document}
\title{Towards Source-free Domain Adaptive Semantic Segmentation via Importance-aware and Prototype-contrast Learning}

\author{Yihong~Cao,
	Hui~Zhang,~\IEEEmembership{Member,~IEEE,}
	Xiao~Lu,
	Zheng~Xiao,
    Kailun~Yang,
	and~Yaonan~Wang%
\thanks{This work was supported in part by the Major Research plan of the National Natural Science Foundation of China under Grant 92148204, the National Natural Science Foundation of China under Grants 62027810 and 61971071, Hunan Leading Talent of Technological Innovation under Grant 2022RC3063, Hunan Science Fund for Distinguished Young Scholars under Grant 2021JJ10025. \textit{(Corresponding author: Hui Zhang.)}}
\thanks{Y. Cao is with the College of Computer Science and Electronic Engineering, Hunan University, Changsha 410082, China, and also with the National Engineering Research Center of Robot Vision Perception and Control Technology, School of Robotics, Hunan University, Changsha 410082, China (email:caoyihong@hnu.edu.cn).}%
\thanks{H. Zhang, K. Yang, and Y. Wang are with the National Engineering Research Center of Robot Vision Perception and Control Technology, School of Robotics, Hunan University, Changsha 410082, China (email: zhanghuihby@126.com, kailun.yang@hnu.edu.cn, yaonan@hnu.edu.cn).}%
\thanks{X. Lu is with the College of Engineering and Design, Hunan Normal University, Changsha 410082, China (e-mail: luxiao@hunnu.edu.cn).}%
\thanks{Z. Xiao is with the College of Computer Science and Electronic Engineering, Hunan University, Changsha 410082, China (e-mail: zxiao@hnu.edu.cn).}%
}

\markboth{IEEE Transactions on Intelligent Vehicles, March~2024}%
{Cao \MakeLowercase{\textit{et al.}}: Towards Source-free Domain Adaptive Semantic Segmentation}

\maketitle

\begin{abstract}
Domain adaptive semantic segmentation enables robust pixel-wise understanding in real-world driving scenes. Source-free domain adaptation, as a more practical technique, addresses the concerns of data privacy and storage limitations in typical unsupervised domain adaptation methods, making it especially relevant in the context of intelligent vehicles. It utilizes a well-trained source model and unlabeled target data to achieve adaptation in the target domain. However, in the absence of source data and target labels, current solutions cannot sufficiently reduce the impact of domain shift and fully leverage the information from the target data. In this paper, we propose an end-to-end source-free domain adaptation semantic segmentation method via \emph{Importance-Aware and Prototype-Contrast (IAPC)} learning. The proposed IAPC framework effectively extracts domain-invariant knowledge from the well-trained source model and learns domain-specific knowledge from the unlabeled target domain. Specifically, considering the problem of domain shift in the prediction of the target domain by the source model, we put forward an importance-aware mechanism for the biased target prediction probability distribution to extract domain-invariant knowledge from the source model. We further introduce a prototype-contrast strategy, which includes a prototype-symmetric cross-entropy loss and a prototype-enhanced cross-entropy loss, to learn target intra-domain knowledge without relying on labels. A comprehensive variety of experiments on two domain adaptive semantic segmentation benchmarks demonstrates that the proposed end-to-end IAPC solution outperforms existing state-of-the-art methods. The source code is publicly available at~\url{https://github.com/yihong-97/Source-free-IAPC}.
\end{abstract}


\IEEEpeerreviewmaketitle

\section{Introduction}

\IEEEPARstart{S}{emantic}
segmentation, a critical visual task in the field of autonomous driving systems, provides pixel-level semantic scene understanding for intelligent vehicles~\cite{Zhang_2022_ExploringEventDriven, Yang_2023_ExploringShape, Fan_2023_MLFNetMultiLevel,zhang2023cmx,qiu2023subclassified}.
\revisedF{Recent years have seen significant advancements in pixel-level segmentation, mainly attributed to the utilization of deep convolutional neural networks~\cite{zhang2019mask, Fan_2023_SegTransConvTransformer, Liu_2023_ImportanceBiased}. }
However, training a network requires a lot of time and labeled data, and it takes $90$ minutes to finely annotate an image of semantic segmentation of high-resolution driving scene images~\cite{Cordts_2016_CityscapesDataset}. 
Furthermore, achieving good performance for a model trained in a specific scene (source domain) in another scene (target domain) is more challenging. \revisedF{This difficulty arises from its limited ability to generalize when encountering previously unseen driving environments within the actual complex and open traffic environment \cite{zhang2021c2fda}.}
All these challenges have emerged as critical bottlenecks in the development of robust and safe autonomous driving systems\revisedF{\cite{zhang2022parallel}}. Unsupervised Domain Adaptation (UDA) is a solution to migrate knowledge from the source domain to the unlabeled target domain through aligned distribution. 
\begin{figure}[!t]
	\centering
	\includegraphics[width=3.5in]{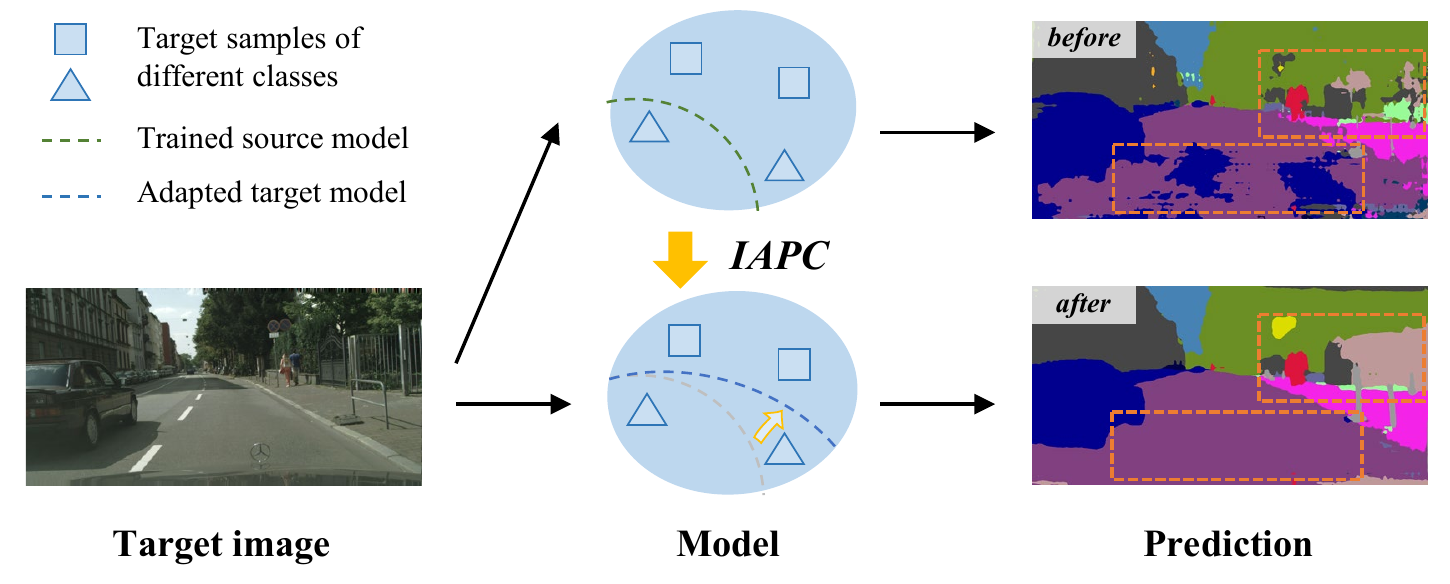}
	\caption{The trained source model shows poor performance when directly making predictions on the target images. With the proposed IAPC framework, a significant improvement is achieved in road-driving scene segmentation.}
	\label{fig0}
\end{figure}

Typical UDA methods~\revisedF{\cite{zhang2021c2fda}}\cite{zhang2021transfer, Hoyer_2023_MICMasked, Ni_2023_SceneAdaptive3D, Michieli_2020_AdversarialLearning} usually require accessing data from both the source and target domain simultaneously, which is restricted when encountering privacy security or storage problems. \revisedF{As mentioned in \cite{zhang2022parallel}, this constraint is especially pertinent in the context of intelligent transportation systems.}
Specifically, the GTA5 datasets~\cite{Lample_2017_PlayingFPS}, commonly used for semantic segmentation, require $57$GB of storage space.
Moreover, it should be noted that due to data privacy concerns, some of the source domain data used for training may not be available for use in practice.
To solve these problems, Source-Free Domain Adaptation (SFDA) techniques~\cite{ Yang_2021_GeneralizedSourceFree, Yin__CrossMatchSourceFreea, Lee_2022_ConfidenceScore, Zhao_2022_SourceFreeOpen, Tian_2022_VDMDAVirtual} are proposed.
\revisedT{The data from the source domain is inaccessible due to either data privacy concerns or storage limitations. Only the model trained on the source domain and unlabeled target domain data can be provided.}
Under this setting, the current UDA methods are not applicable. Specifically, adversarial Learning (AL) based UDA methods~\cite{Michieli_2020_AdversarialLearning, Tian_2022_PartialDomain, Tsai_2018_LearningAdapt} require access to both the source and target domains to align their distributions,
whereas self-training UDA methods~\cite{Zou_2018_UnsupervisedDomain, yang2019recomputation, Lu_2022_BidirectionalSelfTraining} can be prone to catastrophic error growth without guidance from labeled data in the source domain. 

Recently, several SFDA methods for semantic segmentation have been proposed, and they can be roughly categorized into two kinds:
generative methods and self-training-based methods.
Generative SFDA methods~\cite{Liu_2021_SourceFreeDomain, Kundu_2021_GeneralizeThen} generate synthetic source-like data to fit the distribution of the source domain.
Despite their impressive results, generating more reliable source-like data poses a new challenge and brings additional storage resources.
Other SFDA methods~\cite{Akkaya_2022_SelftrainingMetric, Zhao_2023_BetterStability} utilize data augmentation techniques to generate more training samples for the target domain, aiming to address the issues of class imbalance and lack of reliable samples.
\revisedT{The self-training-based SFDA methods~\cite{S_2021_UncertaintyReduction, Yang_2022_SourceFree, Prabhu_2022_AugmentationConsistencyguided, Huang_2021_ModelAdaptation, Karim_2023_CSFDACurriculuma} adopt a more practical and restrictive setting. They rely solely on the well-trained source model and unlabeled target data, adapting through pseudo-labels in the target domain.}
Unprocessed target pseudo-labels generated from the well-trained source model may contain noise. To enhance the effectiveness of pseudo-label learning, one approach~\cite{Yang_2022_SourceFree, Prabhu_2022_AugmentationConsistencyguided, S_2021_UncertaintyReduction, Karim_2023_CSFDACurriculuma} is to employ threshold filtering for low-quality pseudo-labels.
\revised{However, due to the limited number of target labels, these methods can only partially align the knowledge of the source model with the target domain. 
Another strategy involves leveraging prototypes~\cite{Chen_2021_SourceFreeDomain, Huang_2021_ModelAdaptation} to improve the quality of pseudo-label learning. This can be achieved by using prototype-feature similarity as an additional measure of credibility for pseudo-labels~\cite{Chen_2021_SourceFreeDomain} or by employing prototype contrastive learning~\cite{Huang_2021_ModelAdaptation}.
Nevertheless, modeling the prototypes based on the target features still relies on the relevant knowledge extracted from the source model.}

\revisedT{Different from those approaches, we introduce an importance-aware mechanism to learn domain-invariant knowledge from biased target pseudo-labels in the presence of domain shift. Additionally, we combine a prototype-contrast strategy to adapt the model to the target domain by learning specific knowledge.}
This enables our method to effectively address challenges related to the limited number of pseudo-labels and the insufficient learning of target data.

\begin{figure}[!t]
	\centering
	\includegraphics[width=3.5in]{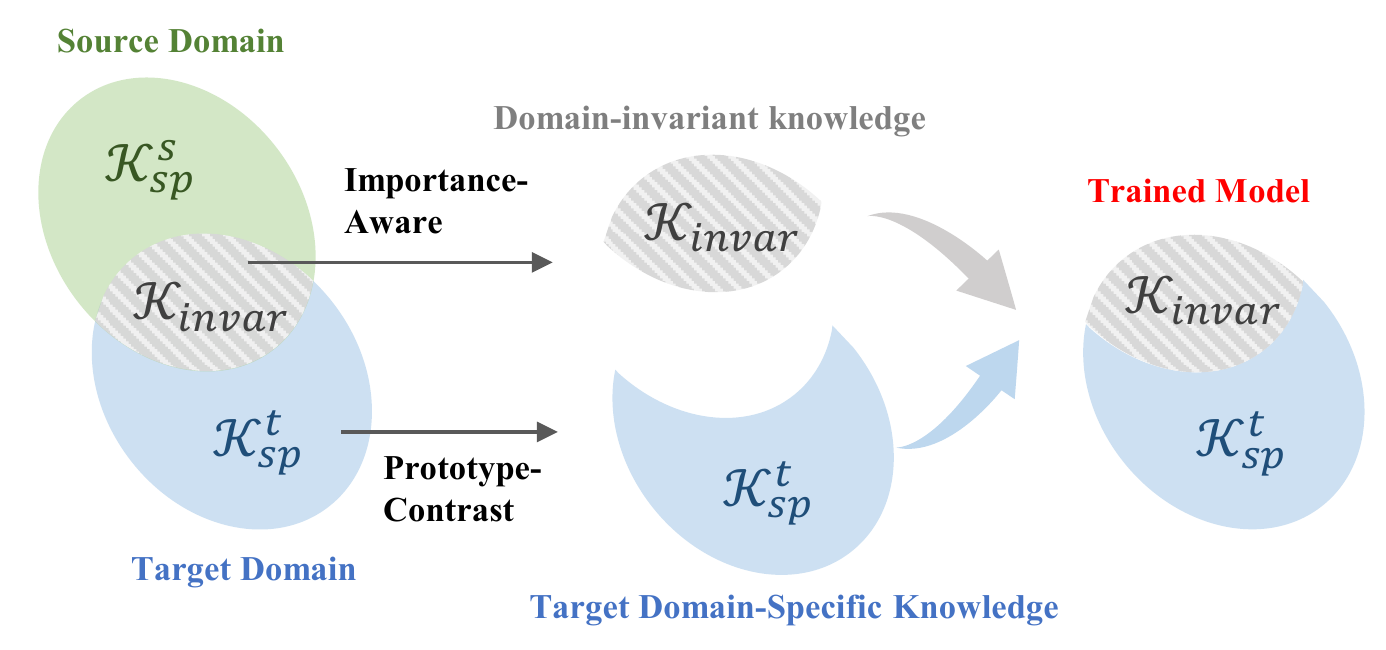}
	\caption{Task of IAPC. The importance-aware mechanism is employed to extract domain-invariant information $\mathcal{K}_{invar}$. Furthermore, the prototype-contrast strategy is utilized to learn target domain-specific knowledge $\mathcal{K}_{sp}^t$. As a result, the trained segmentation model that performs excellently on the target data is obtained.}
	\label{fig1}
\end{figure}
\revised{As shown in Fig.~\ref{fig1}, in the absence of source data and target labels, we propose an \textsc{Importance-Aware and Prototype-Contrast (IAPC)} framework to address this goal from two aspects: }
(1) Extracting Domain-Invariant Knowledge (EDIK) from the trained source model, 
and (2) Learning Domain-Specific Knowledge (LDSK) from the unlabeled target domain.
For EDIK, we observe that the impact of domain shift is directly reflected in the predicted class probability distribution when the well-trained source model predicts the target data.
The source model exhibits higher confidence in predicting data with minimal distribution differences between the source and target domains. Insufficient confidence leads to maximum prediction probability bias towards other categories.
Therefore, we propose an importance-aware mechanism that effectively extracts domain-invariant knowledge from the biased target prediction probability distribution under the influence of domain shift.
For LDSK, we introduce a prototype-contrast strategy to learn domain-specific knowledge in the absence of labeled target data.
The prototypes, estimated by a delayed-updating memory network, serve as anchors and provide reference distribution probabilities and predictions for the target data.
The prototype-symmetric cross-entropy loss and the prototype-enhanced cross-entropy loss are designed to enhance the model's fit to the target task at both the feature and output levels.
Finally, through the synergy of EDIK and LDSK, our model achieves accurate segmentation in the target domain even in the absence of source data and target labels.

At a glance, our contributions are summarized as follows:

\revisedT{(1) We propose the Importance-Aware and Prototype-Contrast (IAPC) framework, an end-to-end solution to source-free domain adaptive semantic segmentation. This framework adapts the capabilities of the well-trained source model to the target domain without accessing source data and target labels.}

(2) We design an importance-aware mechanism to extract domain-invariant knowledge from the well-trained source model and develop a prototype-contrast strategy to learn domain-specific knowledge from the unlabeled target domain.

(3) We evaluate our method on two benchmarks, GTA5$\rightarrow$Cityscapes and SYNTHIA$\rightarrow$Cityscapes, and the results show that our method achieves state-of-the-art results compared to existing SFDA methods.

\section{Related Work}
\subsection{Domain Adaptive Semantic Segmentation}
Domain Adaptive Semantic Segmentation (DASS) is a crucial task in computer vision that aligns the distribution of the source and target domains to improve the model's performance in the unlabeled target domain.
It effectively solves the challenge of time-consuming annotation of training data in target domain scenes.
Existing DASS methods can be roughly categorized into two types: adversarial learning (AL) based methods~\cite{Yu_2021_DASTUnsupervised, zhang2022confidence} and self-training (ST) based methods~\cite{ou2022plume, Spadotto_2021_UnsupervisedDomain, Zheng_2021_RectifyingPseudo}.
\revisedT{AL-based methods train one or multiple domain discriminators to distinguish whether the input comes from the source or target domain. This prompts the segmentation model to generate domain-invariant features to counter the discriminator.}
AL can be applied at the feature level~\cite{Tsai_2018_LearningAdapt, Michieli_2020_AdversarialLearning} and output level~\cite{Hoffman_2016_FCNsWild,zhang2022confidence}.
Although the introduction of AL can effectively align the data distribution of different domains to improve the performance in the target domain, accessibility to the source data is necessary.

ST-based methods generally leverage confident predictions as pseudo-labels to optimize the model.
Considering that the pseudo-labels in the target domain may contain noises, two common methods to address this issue are threshold-based filtering~\cite{Spadotto_2021_UnsupervisedDomain} and uncertainty-guided filtering~\cite{Zheng_2021_RectifyingPseudo, Lu_2022_BidirectionalSelfTraining}.
Traditional DASS methods require simultaneous access to data from both the source and target domains, which is impractical due to data privacy and storage limitations.
Therefore, adapting the model to the target domain using only unlabeled target data without using source domain data is essential.
To this end, we explore domain adaptation without source data and propose a source-free domain adaptive semantic segmentation method via importance-aware and prototype-contrast learning.

\subsection{Source-free Domain Adaptation}

In recent years, source-free domain adaptation techniques that use a well-trained model in the source domain instead of source domain data to adapt to the unlabeled target domain have received increasing attention.
Due to the unavailability of source domain scene data, some generative SFDA methods have been proposed.
Li~\textit{et al.}~\cite{Li_2020_ModelAdaptation} develop a collaborative class conditional generative adversarial networks for producing target-style training samples to improve the prediction model.
\revisedT{Liu~\textit{et al.}~\cite{Liu_2021_SourceFreeDomain} leverage a generator to estimate the source domain and generate fake samples similar to the real source data in distribution. This enables the well-trained source model to transfer the source knowledge.}
\revisedT{Yin~\textit{et al.}~\cite{Yin__CrossMatchSourceFreea} propose a novel asymmetric two-stream architecture for SFDA, achieving excellent performance. They introduce a multimodal auxiliary network that takes depth modality as an additional input.}
\revisedT{While these approaches can enhance the performance of model adaptation, they introduce additional computational overhead or require additional data information.}

\revisedT{Therefore, more SFDA methods comply with the restriction of only using a well-trained model in the source domain and unlabeled training image samples in the target domain. Likewise, our work is conducted under this condition.}
\revised{They typically achieve model adaptation in the target domain through learning target pseudo-labels, and the pseudo-label learning can be roughly divided into prediction-based and prototype-based.}
For prediction-based pseudo labeling methods~\cite{S_2021_UncertaintyReduction, Karim_2023_CSFDACurriculuma, Kothandaraman_2021_SSSFDASelfsupervised}, it is widely considered to filter out noise in predictions by setting the threshold.
In addition to directly using a certain threshold for filtering, adopting the class-balanced threshold strategy to select a certain percentage of predictions based on the threshold set for each class is a more effective approach~\cite{S_2021_UncertaintyReduction, Karim_2023_CSFDACurriculuma}.
\revisedT{Alternatively, Wang~\textit{et al.}~\cite{Wang_2023_CalSFDASourceFreea} propose estimating the expected calibration error (ECE) to eliminate erroneous pseudo-labels with high confidence. The ECE can also serve as an effective generalization indicator for model optimization and selection.}
\revised{Some prototype-based pseudo labeling methods~\cite{ Kim_2021_DomainAdaptation, Yang_2021_TransformerBasedSourceFree, Qiu_2021_SourcefreeDomain, Liang_2021_WeReally} estimate prototypes first and then calculate the similarity between prototypes and the features to obtain pseudo-labels.
\revisedT{Since prototype-based pseudo-labels also contain noise, further using prototype-features similarity~\cite{Hegde_2021_AttentivePrototypes, Qiu_2021_SourcefreeDomain} or prediction consistency~\cite{Chen_2021_SourceFreeDomain, Prabhu_2022_AugmentationConsistencyguided} as the reliability of pseudo labels is an effective way.}
In addition, prototype contrast learning and the memory bank are also widely used in pseudo-label learning.
Huang~\textit{et al.}~\cite{Huang_2021_ModelAdaptation} propose historical contrastive learning to make up for the absence of source data through contrastive learning over embeddings generated from current and historical models. 
Qiu~\textit{et al.}~\cite{Qiu_2021_SourcefreeDomain} introduce weighted prototype contrastive learning to align the target data with source prototypes, and further design a memory bank to regularize early learning. 
Kim~\textit{et al.}~\cite{Kim_2021_DomainAdaptation} design an adaptive prototype memory bank to assign pseudo-labels to target samples.}
\revisedT{In this work, we aim to explore contrastive learning from both prototype and prediction to assist the network in comprehensively learning target domain-specific information. The estimated prototype is derived from the direct output of a delay-updated memory network.}
Furthermore, some methods explore the utilization of style transfer techniques to leverage the knowledge acquired by the source model.
Paul~\textit{et al.}~\cite{Paul_2021_UnsupervisedAdaptation} decrease the shifted distribution between the source and target domains through updating only the normalization parameters of the network with the unlabeled target data.
Zhao~\textit{et al.}~\cite{Zhao_2022_SourceFreeOpen} propose a cross-patch style swap module to diversify samples with various patch styles at the feature level.

In this paper, we propose an importance-aware mechanism to extract domain-invariant knowledge from the pseudo-labels generated by the well-trained source model. Additionally, we introduce a prototype-contrast strategy to learn domain-specific knowledge from the target domain data.

\begin{figure*}[!t]
	\centering
	\includegraphics[width=7in]{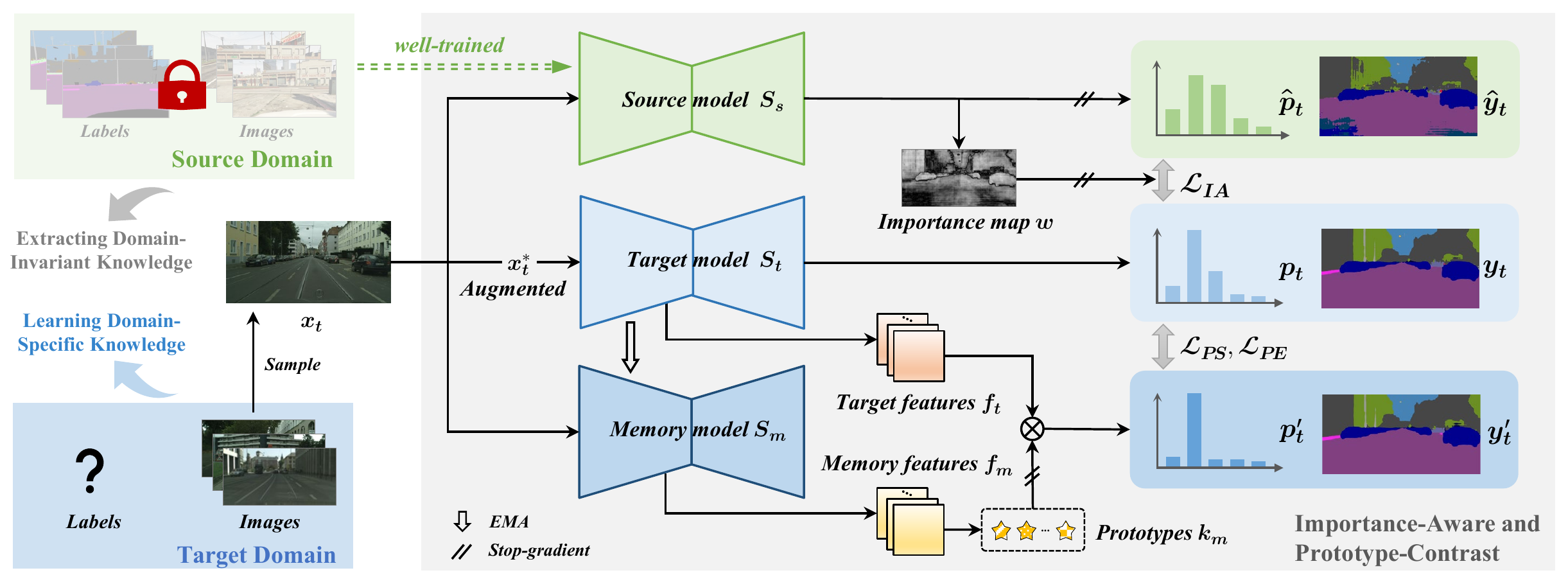}
	\caption{The framework of our proposed IAPC. It effectively addresses the problem of model adaptation when the source domain is inaccessible and the target domain is unlabeled. The augmented image $x^*_t$ is fed into the target model $S_t$, whereas the original image $x_t$ is separately fed into the source model $S_s$ and the memory model $S_m$. Guided by the importance map $w$, the predictions $\hat{p}_t, \hat{y}_t$ from the source model and target predictions $p_t, y$ are combined to extract domain-invariant knowledge under the constraint of $\mathcal{L}_{IA}$. The features $f_m$ from the memory network are used to estimate prototypes, and combined with target features $f_t$ are utilized together by using the prototype-contrast strategy to enable domain-specific knowledge learning within the target domain under the constraint of $\mathcal{L}_{PS}$ and $\mathcal{L}_{PE}$.}
	\label{fig2}
\end{figure*}

\section{IAPC: Proposed Framework}

In the setting of SFDA for semantic segmentation, the task is to adapt the well-trained source model $S_s$ from the source domain $x_s \in \{\mathcal{X}_s\}$ to the unlabeled target domain $x_t \in \{\mathcal{X}_t\}$, where $x_s,x_t \in \mathbb{R}^{H\times W\times 3}$ and $H, W$ represent the height and width of the images.
As shown in Fig.~\ref{fig2}, our proposed IAPC framework consists of two parts: EDIK and LDSK.
As aligning the data distribution between the source and target domains is not feasible, we propose an Importance-Aware (IA) mechanism to enable the training target model $S_t$ to realize domain-invariant knowledge extraction from $S_s$.
Due to the lack of target data labels, we propose a Prototype-Contrast (PC) strategy to facilitate the $S_t$ to complete domain-specific knowledge learning in the target domain.

\subsection{Importance-Aware Learning in EDIK}
For the SFDA task, the information related to the segmentation ability is the well-trained source model, which has excellent performance on the source domain data distribution.
However, due to the domain shift, the performance of the non-adapted model will sharply decline on the target domain.
Therefore, it is necessary to adapt the knowledge acquired by the source model to the target scenario.
Commonly, the well-trained source model $S_s$ is used to generate prediction probability distributions $\hat{p}_t\in \mathbb{R}^{H\times W\times C}$ and pseudo-labels $\hat{y}_t\in \mathbb{R}^{H\times W\times C}$ of the target domain for training, where $C$ denotes the number of classification categories. 
The pseudo-labels $\hat{y}_t$ are one-hot vectors, which are obtained from
\begin{equation}
\label{eq1}
    \hat{y}_t^{h,w} = onehot(\arg\max\limits_{c}\hat{p}_t^{h,w,c}), \text{where~} \hat{p}_t=S_s(x_t),
\end{equation}
where $onehot$ represents the one-hot encoding operation and $(h, w, c)$ represents the index of the position.
However, incorrect predictions in noisy pseudo-labels can lead to poor adaptation performance. To address this issue, we first conduct a theoretical analysis and a practical observation. 

\begin{figure}[!t]
	\centering
	\includegraphics[width=3.5in]{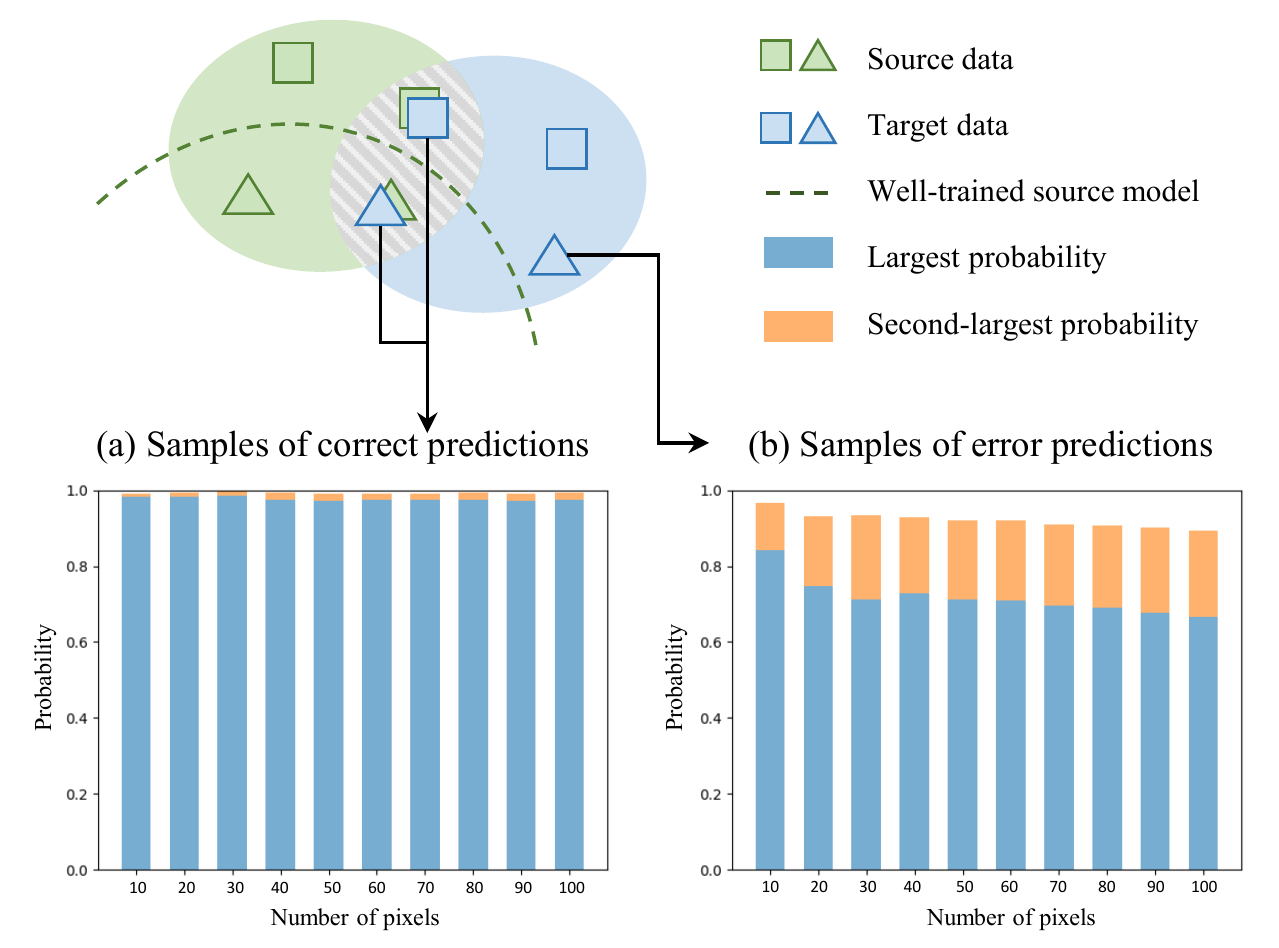}
	\caption{%
\revised{We conducted the pixel-level random sampling of both correct and incorrect predictions by the non-adapted source model on the target data. For each average probability distribution of the selected pixels, we visualized the top two probability values.}
 }
	\label{fig3}
\end{figure}

As shown in Fig.~\ref{fig1}, due to the difference in data distributions, only a portion of the data distributions between the source and target domains overlap. This overlapping portion represents domain-invariant knowledge $\mathcal{K}_{invar}$, while the distinct characteristics of each domain represent domain-specific knowledge $\mathcal{K}_{sp}$.
When the source model $S_s$ is trained on the source domain $\{\mathcal{X}_s\}$, it acquires knowledge that includes both domain-invariant knowledge $\mathcal{K}_{invar}$ and source domain-specific knowledge $\mathcal{K}_{sp}^s$, represented as
\begin{equation}
    \mathcal{F}(S_s) = \mathcal{K}_{invar} + \mathcal{K}_{sp}^s.
\end{equation}
When $S_s$ is directly used to make predictions $\hat{p}_t$ on a target image $x_t$, the shift between the source and target domains leads to expected prediction $p_t = S_t(x_t)$ biases, denoted as
\begin{equation}
\begin{aligned}
    \mathcal{F}(\hat{p}_t)= \mathcal{F}(S_s(x_t))&= \mathcal{K}_{invar}(x_t) + \mathcal{K}_{sp}^s(x_t) \\
    &=\mathcal{F}(S_t(x_t)) - \mathcal{K}_{sp}^t(x_t) + \mathcal{K}_{sp}^s(x_t)\\
    &\approx\mathcal{F}(p_t)+ \varepsilon(\mathcal{K}_{sp}^s,\mathcal{K}_{sp}^t),
\end{aligned}
\end{equation}
where~$\mathcal{F}(S_t) = \mathcal{K}_{invar} + \mathcal{K}_{sp}^t$. The $\mathcal{K}_{sp}^t$ denotes the target domain-specific knowledge. The $\varepsilon(\mathcal{K}_{sp}^s,\mathcal{K}_{sp}^t)$ represents the knowledge bias between the target and source domains, indicating the shift of knowledge between the two domains.
\revisedT{This bias weakens the confidence of the source model in predicting the target domain, which is reflected in the uncertainty of the final predicted class probability distribution $\hat{p}_t$.}
In Fig.~\ref{fig3}, we randomly sample the predicted probability distribution of some pixel positions in a target image and visualize the top two largest probability statistical values of a certain number of pixels in the bar charts.
As shown in Fig.~\ref{fig3}~(a), the source model exhibits greater confidence for correctly predicted pixels, \textit{i.e.}, the model assigns a very high probability to a certain category for that pixel compared to other categories.
Conversely, as shown in Fig.~\ref{fig3}~(b), under the influence of domain-specific knowledge bias, the model tends to exhibit prediction biases for pixels with incorrect predictions, which is reflected by a relatively large value for the second-largest probability.
That is, the difference between the probability values of the largest probability and the second-largest probability is positively correlated with the bias $\varepsilon$ between the domains. 

We first design a simple yet effective method to estimate the importance $\omega^{h,w}$ of each pixel position, denoted as
\begin{equation}
    \omega^{h,w} = 1- \frac{\max\limits_{c\in C, c\neq \hat{y}_t^{h,w}}{\hat{p}_t^{h,w,c}}}{\max\limits_{c\in C}{\hat{p}_t^{h,w,c}}}.
\end{equation}
If the source model is confident about the prediction of the current position, the importance value will be close to $1$.
Conversely, the uncertainty of the model prediction will lead to a lower importance value. The IA-based pseudo-label loss $\mathcal{L}_{IA}$ is given by
\begin{equation}
    \mathcal{L}_{IA} =  - \frac{1}{HW}\sum\limits_{h,w}\omega^{h,w}\sum\limits_{c}{\hat{y}_t^{h,w,c}\log {p_t^{h,w,c}}}.
\end{equation}

The $\mathcal{L}_{IA}$ effectively knowledge of the small shift in the source model to the target model under the guidance of the importance map. Furthermore, we argue the target probability distribution should be close to a one-hot encoding. To achieve this, we adopt the information maximization loss $\mathcal{L}_{IM}$ to enhance the certainty and diversity of the target predictions.
\begin{equation}
    \mathcal{L}_{IM} =  - \frac{1}{HW}\sum\limits_{h,w}\sum\limits_{c}{p_t^{h,w,c}\log{p_t^{h,w,c}}}.
\end{equation}

The combination of $\mathcal{L}_{IA}$ and $\mathcal{L}_{IM}$ effectively enables the transfer of domain-invariant knowledge from the source domain to the target domain when only the well-trained source model is available for access.

\subsection{Prototype-contrast Learning in LDSK}
Extracting domain-invariant knowledge from the source model can only ensure that the model has the segmentation ability for common information (\textit{i.e.}, similar distributions between the source and target domains).
In order to further improve the model's segmentation ability for the target domain, we need to further excavate the proprietary information within the target domain.

Considering that feature prototypes are less sensitive to the minority outliers and irrelevant to the frequency of category occurrence, we propose a prototype contrast strategy to learn the target domain-specific knowledge.
It utilizes a delay-updated memory model $S_m$ to calculate feature prototypes $k_m$ and further guide the model's self-adaptation. 
Following~\cite{Yang_2021_TransformerBasedSourceFree, Huang_2021_ModelAdaptation, Karim_2023_CSFDACurriculuma}, the memory model $S_m$ uses the parameters of the target model $S_t$ to update with the exponential moving average (EMA).
\revised{This processing involves smoothing historical information to alleviate instability in the training process and prevent interference from noise in the training data. It enables the prototype-contrast learning strategy to more accurately capture the dynamic changes in the target data during the current iteration, facilitating better learning of domain-specific knowledge.}
\revised{The input image $x_t$ and augmented image $x^*_t$ are fed into the $S_m$ and $S_t$ respectively to obtain intermediate features $f_m,~f_t$ and predictions $p_m,~p_t$.}
\revised{
\begin{equation}
\begin{split}
    f_m = F_m(x_t),&~p_m = S_m(x_t),\\
    f_t = F_t(x^*_t),&~p_t = S_t(x^*_t),
\end{split}
\end{equation}
}\revised{where the $F_m$ and $ F_t$ represent the feature extraction encoders within $S_m$ and $S_t$, respectively.}
\revised{Considering that calculated prototypes in the hidden feature space are prone to be suboptimal, we adopt a prototype calculation strategy that differs from existing methods~\cite{Yang_2021_TransformerBasedSourceFree, Qiu_2021_SourcefreeDomain, Kim_2021_DomainAdaptation} for SFDA in semantic segmentation.}
On the one hand, the features $f_m$ used for prototype calculation are obtained by the delay-update memory model $S_m$.
\revised{This choice ensures the stability of prototype estimation, providing a reliable basis for domain-specific learning.}
On the other hand, a separate prototype is estimated for each input image $x_t$, rather than using the features of a batch or all samples. Since semantic segmentation is a pixel-level classification task, we believe that using the feature from a single image is sufficient for prototype estimation. 
\revisedT{Contrarily, static or iteratively smoothed prototype approaches diverge from our objective of conducting local contrast learning for the current target image. They jeopardize the following two losses designed to enhance prediction consistency.}
Our dynamic and independent prototypes aim to capture intricate details in the image, facilitating the model in more effectively extracting domain-specific knowledge from the current target image.
\revised{Therefore, given the features $f_m$ and predictions $p_m$ from the memory model $S_m$, the prototype $k_m^c$ of class $c$ can be obtained via the following equation:}
\begin{equation}
	{{k}_m^c} = \frac{\sum\limits_{h,w}y_m^{h,w,c}f_m^{h,w}}{\sum\limits_{h,w}y_m^{h,w,c}}, c \in {C},
	\label{eq_3}
\end{equation}
\revised{where the pseudo-labels $y_m^{h,w}$ are the one-hot vectors based on the predictions $p_m$ and are obtained from}
\begin{equation}
\revised{
    {y}_m^{h,w} = onehot(\arg\max\limits_{c}{p}_m^{h,w,c}).}%
\end{equation}
\revised{Then, we can calculate the similarity between the features $f_t$ output by the target network and each prototype $k_m^c$ based on the memory model to obtain the reference class probability distribution ${p'}_t^{h,w,c}$ and the reference pseudo-labels $y'_t$:}%
\begin{equation}
	{{p'}_t^{h,w,c}} = \frac{f_t^{h,w}\cdot {{{k}}_m^c}^T}{\sum\limits_{c}f_t^{h,w}\cdot {{{k}}_m^c}^T}.
	\label{eq_4}
\end{equation}
\begin{equation}
\revised{
    {y'}_t^{h,w} = onehot(\arg\max\limits_{c}{p'}_t^{h,w,c}).}
\end{equation}
\revised{Specifically, by using the features $f_m, f_t$ respectively from the memory model and the target model to calculate $p'_t$ and $y'_t$, we can avoid the association between the prototype and individual features, and improve the accuracy of the prototype-based class probability distribution.}
\revised{To this end, distinct from existing prototype contrast learning techniques~\cite{Qiu_2021_SourcefreeDomain, Huang_2021_ModelAdaptation}, we propose a novel prototype contrast strategy combined with prototype- and prediction-based pseudo labels. We aim to learn domain-specific knowledge using accessible target data guided by target-relevant models $S_t, S_m$.}
We first design the following prototype-symmetric cross-entropy loss $\mathcal{L}_{PS}$ to promote consistency in the probability distribution of model predictions at both the instance and category levels:
\begin{equation}
\begin{aligned}
    \mathcal{L}_{PS} =  &- \frac{1}{HW}\sum\limits_{h,w}\sum\limits_{c}({{y'}_t^{h,w,c}\log{p_t^{h,w,c}}}\\
    &+{{y_t}^{h,w,c}\log{{p'}_t^{h,w,c}}}).
\end{aligned}
\end{equation}
Subsequently, we design a prototype-enhanced cross-entropy loss $\mathcal{L}_{PE}$, which leverages prototype-based reference predictions ${y'}_t$ to enhance the target model's performance on hard classes: 
\begin{equation}
    \mathcal{L}_{PE} =  - \frac{1}{HW}\sum\limits_{h,w}\sum\limits_{c}{{\mathbbm{1}_{[y_t^{h,w,c}={y'}_t^{h,w,c}]}}{y'}_t^{h,w,c}\log{p_t^{h,w,c}}}.
\end{equation}
Compared to the $\mathcal{L}_{IA}$ in EDIK, dynamically updated prototypes during the iteration process can better utilize information within the target domain to guide the model adaptation.

\revisedT{At this stage, we leverage a delayed update memory model to guide the estimation of the dynamic prototypes. Then, we promote the compression of intra-class features and the separation of inter-class features associated with the proposed $\mathcal{L}_{PS}$ and $\mathcal{L}_{PE}$.}
\revised{%
This effectively enhances the target model's acquisition of domain-specific information, thereby improving its adaptation capability.}

\begin{table*}[htbp]
	\centering
	\caption{Experimental results of the \textit{GTA5 $\rightarrow$ Cityscapes} adaptation scene in terms of per-category IoU, mIoU. The \textbf{SF} column indicates whether the adaptive method is source-free. The best results are shown in bold.}
	\setlength{\tabcolsep}{0.8mm}{
		\begin{tabular}{c|c|p{6.0mm}<{\centering}p{6.0mm}<{\centering}p{6.0mm}<{\centering}  p{6.0mm}<{\centering}   p{6.0mm}<{\centering}p{6.0mm}<{\centering}p{6.0mm}<{\centering}p{6.0mm}<{\centering}p{6.0mm}<{\centering}p{6.0mm}<{\centering}p{6.0mm}<{\centering}p{6.0mm}<{\centering}p{6.0mm}<{\centering}p{6.0mm}<{\centering}p{6.0mm}<{\centering}p{6.0mm}<{\centering}p{6.0mm}<{\centering}p{6.0mm}<{\centering}p{6.0mm}<{\centering}|c}
			\toprule
			\multicolumn{1}{c}{Methods} &\multicolumn{1}{c}{\textbf{SF}}& \rotatebox{45}{Road}  & \rotatebox{45}{Sidewalk} & \rotatebox{45}{Building} & \rotatebox{45}{Wall}  & \rotatebox{45}{Fence} & \rotatebox{45}{Pole}  &\rotatebox{45}{Lights} & \rotatebox{45}{Sign}  & \rotatebox{45}{Vegetation} & \rotatebox{45}{Terrain} & \rotatebox{45}{Sky}   & \rotatebox{45}{Person} & \rotatebox{45}{Rider} & \rotatebox{45}{Car}   & \rotatebox{45}{Truck} & \rotatebox{45}{Bus}   & \rotatebox{45}{Train} & \rotatebox{45}{Motorbike} & \multicolumn{1}{c}{\rotatebox{45}{Bicycle}} & mIoU  \\
			\midrule
    Source only &\multirow{6}{*}{\XSolidBrush}& 78.9  & 15.5  & 75.0  & 17.6  & 15.8  & 24.8  & 30.1  & 21.4  & 73.9  & 20.1  & 77.4  & 55.6  & 24.4  & 70.9  & 27.8  & 7.1   & 2.9   & 25.3  & 31.3  & 36.6  \\
    AdaptSegNet \cite{Tsai_2018_LearningAdapt} && 86.5  & 36.0  & 79.9  & 23.4  & 23.3  & 23.7  & 35.2  & 14.8  & 83.4  & 33.3  & 75.6  & 58.5  & 27.6  & 73.7  & 32.5  & 35.4  & 3.9   & 30.1  & 28.1  & 42.4  \\
    CBST \cite{Zou_2018_UnsupervisedDomain} & & 91.8  & 53.5  & 80.5  & 32.7  & 21.0  & 34.0  & 28.9  & 20.4  & 83.9  & 34.2  & 80.9  & 53.1  & 24.0  & 82.7  & 30.3  & 35.9  & \textbf{16.0} & 25.9  & 42.8  & 45.9  \\
    AdvEnt \cite{Vu_2019_ADVENTAdversarial}& & 89.4  & 33.1  & 81.0  & 26.6  & 26.8  & 27.2  & 33.5  & 24.7  & 83.9  & 36.7  & 78.8  & 58.7  & \textbf{30.5} & 84.8 & 38.5  & 44.5  & 1.7   & 31.6  & 32.4  & 45.5  \\
    MaxSquare \cite{Chen_2019_DomainAdaptation}&& 89.3  & 40.5  & 81.2  & 29.0  & 20.4  & 25.6  & 34.4  & 19.0  & 83.6  & 34.4  & 76.5  & 59.2  & 27.4  & 83.8  & 38.4  & 43.6  & 7.1   & 32.2  & 32.5  & 45.2  \\
    UDAClu \cite{Toldo_2021_UnsupervisedDomain} && 89.4  & 30.7  & 82.1  & 23.0  & 22.0  & 29.2  & 37.6  & 31.7  & 83.9  & 37.9  & 78.3  & 60.7  & 27.4  & 84.6  & 37.6  & 44.7  & 7.3   & 26.0  & 38.9  & 45.9  \\
    \midrule
    UR \cite{S_2021_UncertaintyReduction} &\multirow{8}{*}{\Checkmark}& \textbf{92.3} & \textbf{55.2} & 81.6  & 30.8  & 18.8  & \textbf{37.1} & 17.7  & 12.1  & 84.2  & 35.9  & 83.8  & 57.7  & 24.1  & 81.7  & 27.5  & 44.3  & 6.9   & 24.1  & 40.4  & 45.1  \\
    SFDA \cite{Liu_2021_SourceFreeDomain} && 91.7  & 52.7  & 82.2  & 28.7  & 20.3  & 36.5  & 30.6  & 23.6  & 81.7  & 35.6  & 84.8  & 59.5  & 22.6  & 83.4  & 29.6  & 32.4  & 11.8  & 23.8  & 39.6  & 45.8  \\
    LD \cite{You_2021_DomainAdaptive} && 91.6  & 53.2  & 80.6  & \textbf{36.6} & 14.2  & 26.4  & 31.6  & 22.7  & 83.1  & 42.1 & 79.3  & 57.3  & 26.6  & 82.1  & \textbf{41.0} & \textbf{50.1} & 0.3   & 25.9  & 19.5  & 45.5  \\
    HCL \cite{Huang_2021_ModelAdaptation} && 92.0  & 55.0  & 84.0  & 33.5  & 24.6  & \textbf{37.1} & 35.1  & 28.8  & 83.0  & 37.6  & 82.3  & 59.4  & 27.6  & 83.6  & 32.3  & 36.6  & 14.1  & 28.7  & 43.0  & 48.1  \\
    DTAC \cite{Yang_2022_SourceFree} && 78.0  & 29.5  & 83.0  & 29.3  & 21.0  & 31.8  & 38.1  & 33.1  & 83.8  & 39.2  & 80.8  & \textbf{61.0} & 30.3  & 83.9  & 26.1  & 40.4  & 1.9   & \textbf{34.2} & 43.7  & 45.7  \\
    C-SFDA \cite{Karim_2023_CSFDACurriculuma} & & 90.4  & 42.2  & 83.2  & 34.0  & 29.3  & 34.5  & 36.1  & 38.4  & 84.0  & \revisedT{43.0} & 75.6  & 60.2  & 28.4  & \textbf{85.2} & 33.1  & 46.4  & 3.5   & 28.2  & 44.8  & 48.3  \\
    \revisedT{Cal-SFDA\cite{Wang_2023_CalSFDASourceFreea}}& & \revisedT{90.0}  & \revisedT{48.4}  & \revisedT{83.2}  & \revisedT{35.5}  & \revisedT{23.6}  & \revisedT{30.8}  & \revisedT{39.6}  & \revisedT{35.9}  & \revisedT{\textbf{84.3}} & \revisedT{\textbf{43.2}} & \revisedT{85.1}  & \revisedT{60.2}  & \revisedT{27.9}  & \revisedT{84.3}  & \revisedT{32.6}  & \revisedT{44.7}  & \revisedT{2.2}   & \revisedT{19.9}  & \revisedT{42.1}  & \revisedT{48.1}  \\
    IAPC (Ours)  && 90.9  & 36.5  & \textbf{84.4} & 36.1  & \textbf{31.3} & 32.9  & \textbf{39.9} & \textbf{38.7} & \textbf{84.3} & 38.6  & \textbf{87.5} & 58.6  & 28.8  & 84.3  & 33.8  & 49.5  & 0.0   & 34.1  & \textbf{47.6} & \textbf{49.4} \\
    \bottomrule
		\end{tabular}%
	}
	\label{tab1}%
\end{table*}%

\section{Experiments}
\subsection{Datasets}
To facilitate the evaluation and fair comparison of the effectiveness of our IAPC framework, we conduct evaluations on two widely-used cross-domain benchmarks in \textit{GTA5 $\rightarrow$ Cityscapes} and \textit{SYNTHIA $\rightarrow$ Cityscapes}.
GTA5~\cite{Lample_2017_PlayingFPS} contains $24,966$ fully annotated urban scene synthetic images.
The pixel-level ground truth labels are automatically generated by computer graphics and compatible with the format of Cityscapes \cite{Cordts_2016_CityscapesDataset}. We used the $19$ common classes with Citysacpes for training.
SYNTHIA~\cite{Ros_2016_SynthiaDataset} is another synthetic dataset, and commonly the SYNTHIA-RAND-CITYSCAPES subset is used, which contains $9,400$ images.
The ground-truth labels are also compatible with Cityscapes and we select $16$ common classes for training.
Cityscapes~\cite{Cordts_2016_CityscapesDataset} is a real-world dataset acquired while driving in European cities.
It consists of $2,975$ images from the training set and $500$ images from the validation set.
These images are precisely labeled for $19$ semantic classes.

\subsection{Implementation Details}
We adopt the DeepLabV2 architecture~\cite{Chen_2018_DeepLabSemantica} with ResNet-101~\cite{He_2016_DeepResidual} as the segmentation network.
The same architecture is used for the memory network.
For the well-trained source model, following~\cite{Chen_2019_DomainAdaptation, Toldo_2021_UnsupervisedDomain}, we adopt the stochastic gradient descent algorithm with momentum $0.9$, weight decay $5e{-}4$ and learning rate $2.5e{-}4$.
For the source-free target adaptation stage, the learning rate is set as $1e{-}4$.
For both stages, the decrease in learning rate is set at a polynomial decay with a power of $0.9$.
The size of input images is resized to $1024{\times}512$ and cropped $512{\times}256$ patch randomly for training.
Following~\cite{Huang_2021_ModelAdaptation}, we adopt the partial image augmentation in~\cite{Cubuk_2019_AutoaugmentLearning} to perform photometric noise on the input data of the target network.
The memory network is updated with the parameters of the target network with the smoothing factor $1e{-}4$.
As for the hyper-parameters, the weights of the four losses $\mathcal{L}_{IA},\mathcal{L}_{PE},\mathcal{L}_{PS}$, and $\mathcal{L}_{IM}$ are set to $0.2$, $0.5$, $0.01$, and $2$ on the two benchmarks. 
The batch size is set to $6$.
The proposed IAPC is implemented with the PyTorch toolbox on a single NVIDIA GeForce RTX 3090 GPU.

The metric of pixel-wise semantic segmentation performance is measured by the Intersection over Union (IoU) of each category.
The mean IoU of all training categories is used to compare the overall performance of the methods.
Following previous methods, mIoU* ($13$ common categories in SYNTHIA excluding wall, fence, and pole) is added in the benchmark of \textit{SYNTHIA $\rightarrow$ Cityscapes}.

\begin{figure*}[!t]
	\centering
	\includegraphics[width=7in]{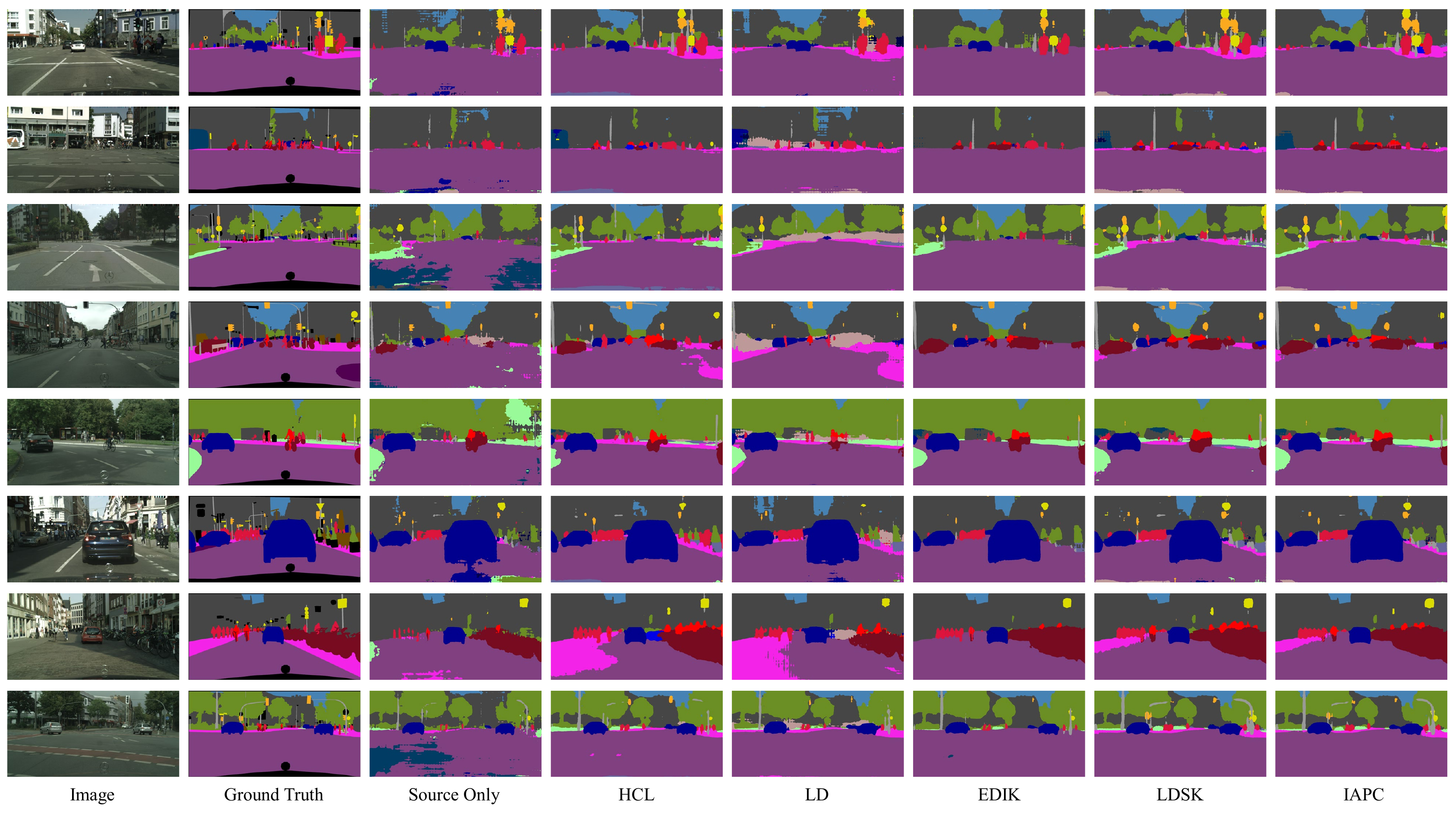}
	\caption{Qualitative results of semantic segmentation adaptation on GTA5 $\rightarrow$ Cityscapes. (From top to bottom: Image, Ground Truth, Source Only, HCL~\cite{Huang_2021_ModelAdaptation}, LD~\cite{You_2021_DomainAdaptive}, IAPC-EDIK, IAPC-LDSK, and IAPC results (best viewed in color).}
	\label{fig4}
\end{figure*}

\begin{table*}[t!]
	\centering
	\caption{Experimental results of the SYNTHIA $\rightarrow$ Cityscapes adaptation scene in terms of per-category IoU, mIoU, and mIoU*. The \textbf{SF} column indicates whether the adaptive method is source-free. The best results are shown in bold.}
	\setlength{\tabcolsep}{1.2mm}{
		\begin{tabular}{c|c|p{6.0mm}<{\centering}p{6.0mm}<{\centering}p{6.0mm}<{\centering}  p{6.0mm}<{\centering}   p{6.0mm}<{\centering}p{6.0mm}<{\centering}p{6.0mm}<{\centering}p{6.0mm}<{\centering}p{6.0mm}<{\centering}p{6.0mm}<{\centering}p{6.0mm}<{\centering}p{6.0mm}<{\centering}p{6.0mm}<{\centering}p{6.0mm}<{\centering}p{6.0mm}<{\centering}p{6.0mm}<{\centering}|cc}
			
			\toprule
			\multicolumn{1}{c}{Methods} &\multicolumn{1}{c}{\textbf{SF}}& \rotatebox{45}{Road}  & \rotatebox{45}{Sidewalk} & \rotatebox{45}{Building} & \rotatebox{45}{Wall}  & \rotatebox{45}{Fence} & \rotatebox{45}{Pole}  &\rotatebox{45}{Lights} & \rotatebox{45}{Sign}  & \rotatebox{45}{Vegetation}  & \rotatebox{45}{Sky}   & \rotatebox{45}{Person} & \rotatebox{45}{Rider} & \rotatebox{45}{Car}   &  \rotatebox{45}{Bus}    & \rotatebox{45}{Motorbike} & \multicolumn{1}{c}{\rotatebox{45}{Bicycle}} & mIoU  & mIoU* \\
			\midrule		
Source only &\multirow{6}{*}{\XSolidBrush}& 39.5  & 18.1  & 75.5  & 10.5  & 0.1   & 26.3  & 9.0   & 11.7  & 78.6  & 81.6  & 57.7  & 21.0  & 59.9  & 30.1  & 15.7  & 28.2  & 35.2  & 40.5  \\
AdaptSegNet \cite{Tsai_2018_LearningAdapt}&& 84.3  & \textbf{42.7} & 77.5  &    -  &    -   &  -     & 4.7   & 7.0   & 77.9  & 82.5  & 54.3  & 21.0  & 72.3  & 32.2  & 18.9  & 32.3  &   -    & 46.7  \\
CBST \cite{Zou_2018_UnsupervisedDomain} && 68.0  & 29.9  & 76.3  & 10.8  & 1.4   & 33.9  & 22.8  & 29.5  & 77.6  & 78.3  & \textbf{60.6}  & \textbf{28.3} & 81.6  & 23.5  & 18.8  & 39.8  & 42.6  & 48.9  \\
AdvEnt \cite{Vu_2019_ADVENTAdversarial}& &85.6  & 42.2  & 79.7  & 8.7   & 0.4   & 25.9  & 5.4   & 8.1   & 80.4  & 84.1  & 57.9  & 23.8  & 73.3  & 36.4  & 14.2  & 33.0  & 41.2  & 48.0  \\
MaxSquare \cite{Chen_2019_DomainAdaptation} && 78.5  & 34.7  & 76.3  & 6.5   & 0.1   & 30.4  & 12.4  & 12.2  & 82.2  & 84.3  & 59.9  & 17.9  & 80.6  & 24.1  & 15.2  & 31.2  & 40.4  & 46.9  \\
UDAClu \cite{Toldo_2021_UnsupervisedDomain} && \textbf{88.3} & 42.2  & 79.1  & 7.1   & 0.2   & 24.4  & 16.8  & 16.5  & 80.0  & 84.3  & 56.2  & 15.0  & \textbf{83.5} & 27.2  & 6.3   & 30.7  & 41.1  & 48.2  \\
\midrule
UR \cite{S_2021_UncertaintyReduction} & \multirow{8}{*}{\Checkmark}& 59.3  & 24.6  & 77.0  & \textbf{14.0} & \textbf{1.8} & 31.5  & 18.3  & \textbf{32.0} & \textbf{83.1} & 80.4  & 46.3  & 17.8  & 76.7  & 17.0  & 18.5  & 34.6  & 39.6  & 45.0  \\
SFDA \cite{Liu_2021_SourceFreeDomain} && 67.8  & 31.9  & 77.1  & 8.3   & 1.1   & 35.9  & 21.2  & 26.7  & 79.8  & 79.4  & 58.8  & 27.3  & 80.4  & 25.3  & 19.5  & 37.4  & 42.4  & 48.7  \\
LD \cite{You_2021_DomainAdaptive} & &77.1  & 33.4  & 79.4  & 5.8   & 0.5   & 23.7  & 5.2   & 13.0  & 81.8  & 78.3  & 56.1  & 21.6  & 80.3  & \textbf{49.6} & 28.0  & \textbf{48.1} & 42.6  & 50.1  \\
HCL \cite{Huang_2021_ModelAdaptation} && 80.9  & 34.9  & 76.7  & 6.6   & 0.2   & \textbf{36.1} & 20.1  & 28.2  & 79.1  & 83.1  & 55.6  & 25.6  & 78.8  & 32.7  & 24.1  & 32.7  & 43.5  & 50.2  \\
DTAC \cite{Yang_2022_SourceFree} && 77.5  & 37.4  & 80.5  & 13.5  & 1.7   & 30.5  & \textbf{24.8} & 19.7  & 79.1  & 83.0  & 49.1  & 20.8  & 76.2  & 12.1  & 16.5  & 46.1  & 41.8  & 47.9  \\
C-SFDA \cite{Karim_2023_CSFDACurriculuma} && 87.0  & 39.0  & 79.5  & 12.2  & \textbf{1.8 } & 32.2  & 20.4  & 24.3  & 79.5  & 82.2  & 51.5  & 24.5  & 78.7  & 31.5  & 21.3  & 47.9  & 44.6  & 51.3  \\
\revisedT{Cal-SFDA \cite{Wang_2023_CalSFDASourceFreea}} & & \revisedT{76.3}  & \revisedT{32.6}  & \revisedT{81.2}  & \revisedT{4.0}   & \revisedT{0.6}   & \revisedT{27.5}  & \revisedT{20.2}  & \revisedT{17.6}  & \revisedT{82.4}  & \revisedT{83.1}  & \revisedT{51.8}  & \revisedT{18.1}  & \revisedT{83.3}  & \revisedT{46.2}  & \revisedT{14.7}  & \revisedT{\textbf{48.1}} & \revisedT{43.0}  & \revisedT{50.4}  \\
IAPC (Ours)  & & 68.5  & 29.2  & \textbf{82.0} & 10.9  & 1.2   & 28.7  & 22.3  & 29.1  & 82.8  & \textbf{85.3} & 60.5  & 19.3  & 83.1  & 42.5  & \textbf{32.2} & 47.7  & \textbf{45.3} & \textbf{52.7} \\
			\bottomrule	
		\end{tabular}%
	}
	\label{tab2}%
\end{table*}%

\subsection{Comparison against the State of the Art}
In this part, we present the experimental results of the proposed IAPC on two benchmarks and compare it with other existing SFDA methods~\cite{S_2021_UncertaintyReduction, Huang_2021_ModelAdaptation, Liu_2021_SourceFreeDomain, You_2021_DomainAdaptive, Yang_2022_SourceFree, Karim_2023_CSFDACurriculuma}\revisedT{\cite{Wang_2023_CalSFDASourceFreea}} for semantic segmentation.
We also report results of source data accessible UDA methods~\cite{Tsai_2018_LearningAdapt, Vu_2019_ADVENTAdversarial, Zou_2018_UnsupervisedDomain, Chen_2019_DomainAdaptation, Toldo_2020_UnsupervisedDomain}.
For a fair comparison, all methods adopt the DeepLabV2 architecture~\cite{Chen_2018_DeepLabSemantica} with a ResNet-101 backbone~\cite{He_2016_DeepResidual}.
The results show that IAPC outperforms existing methods and achieves state-of-the-art performance on both benchmarks, demonstrating its effectiveness and progressiveness.

\subsubsection{GTA5 $\rightarrow$ Cityscapes}
Table~\ref{tab1} shows the comparison of the proposed IAPC's performance compared with other methods.
IAPC achieves the best mIoU of $49.4$ without accessing source domain data.
With EDIK and LDSK, IAPC improves the performance of the ``Source Only'' by $35\%$.
This indicates that our proposed strategy can effectively utilize the well-trained source model and unlabeled target domain data to achieve model adaptation in the target domain.
Although the contrastive-learning-based HCL~\cite{Huang_2021_ModelAdaptation} also utilizes information on feature level and prediction level, our IAPC exceeds it by a large gap of $2.7\%$.
This confirms the effectiveness of our proposed IA mechanism for extracting domain-invariant knowledge.
The proposed IAPC outperforms the prediction-based C-SFDA~\cite{Karim_2023_CSFDACurriculuma} by $2.2\%$.
\revisedT{Additionally, IAPC exhibits a $2.7\%$ improvement over Cal-SFDA~\cite{Wang_2023_CalSFDASourceFreea}, which considers further filtering out erroneous pseudo-labels with high confidence. These results further indicate that the IA mechanism effectively suppresses noisy predictions in the absence of target labels.}
\revisedT{Compared to the UDA methods~\cite{Vu_2019_ADVENTAdversarial, Zou_2018_UnsupervisedDomain, Chen_2019_DomainAdaptation, Toldo_2020_UnsupervisedDomain}, IAPC achieves excellent segmentation performance even in the absence of source data. This effectively addresses the concerns about data privacy and capacity limitations in domain adaptive semantic segmentation tasks.}
Fig.~\ref{fig4} visualizes the segmentation results of a set of representative samples.
\revisedT{It can be clearly observed that IAPC effectively completes the adaptation of the source model to the target domain compared to the baseline of ``Source Only''.}

\subsubsection{SYNTHIA $\rightarrow$ Cityscapes}
Table~\ref{tab2} displays the comparison results in another scenario.
Following the same evaluation protocols, we report the experimental results in two metrics: mIoU and mIoU*.
IAPC surpasses existing methods by a significant margin, achieving a surprising mIoU of $45.3$ and mIoU* of $52.6$.
This further confirms the effectiveness of IAPC.
IAPC outperforms HCL~\cite{Huang_2021_ModelAdaptation} by $4.1\%$ in mIoU and $4.7\%$ in mIoU*. Our approach also performs exceptionally well compared to source-accessible UDA methods. UDAClu~\cite{Toldo_2020_UnsupervisedDomain} is a feature-level method that utilizes estimated prototypes for clustering and separation.
In comparison, IAPC gains a $10.2\%$ improvement in mIoU and a $9.1\%$ improvement in mIoU* by employing the feature-level PC strategy. Unlike UDAClu, which can access source domain data to provide more accurate estimates of prototypes, IAPC's prototype estimation does not have access to any real labels. The outstanding performance confirms the effectiveness of the single image prototype dynamic estimation guided by the memory network in the PC strategy.
We observe significant improvements in two categories: \emph{Bus} and \emph{Motorbiker}, which are often easily confused and misclassified by the segmentation model.
This indicates that utilizing the PC in the LDSK strategy can better guide the model to learn the features of these challenging less-frequent categories in the target domain.
\begin{table}[h]
\renewcommand{\arraystretch}{1.2}
  \centering
   \caption{\revised{Effect of different components in our IAPC framework on the GTA5 $\rightarrow$ Cityscapes adaptation scenario.} The best results are shown in bold.}
  \setlength{\tabcolsep}{3mm}{
    \begin{tabular}{c|cccc|c}
    \toprule
     &$\mathcal{L}_{IA}$     & $\mathcal{L}_{IM}  $   & $\mathcal{L}_{PS}$ & $\mathcal{L}_{PE}$  & mIoU \\

          \hline

          \multirow{2}[0]{*}{EDIK}&\Checkmark &       &       &       & 44.1 \\
          &\Checkmark & \Checkmark  &      &       & 46.3 \\
          \hline
           \multirow{3}[0]{*}{LDSK}& &     & \Checkmark &       & 42.2 \\
          & &      &       & \Checkmark & 47.7 \\
          & &    &\Checkmark   &\Checkmark  & 48.1\\
          \hline
          Source only&&       &       &       & 36.6 \\
          IAPC (Ours)&\Checkmark & \Checkmark & \Checkmark & \Checkmark & \textbf{49.4} \\
          \bottomrule
    \end{tabular}%
    }
  \label{tab3}%
\end{table}%

\begin{table}[h]
\renewcommand{\arraystretch}{1.3}
  \centering
   \caption{\revised{Ablation study on the GTA5 $\rightarrow$ Cityscapes adaptation scenario. The $\downarrow$ represents the corresponding gap.}}
  \setlength{\tabcolsep}{1.6mm}{
    \begin{tabular}{c|cc|cc}
    \toprule
        &\revisedT{~~~~~$\mathcal{L}_{IA}$~~~~~}     &  \revisedT{$\mathcal{L}_{PS}+\mathcal{L}_{PE}$ } & \revisedT{mIoU}&\revisedT{$\downarrow$} \\

          \hline

          \multirow{4}[0]{*}{\revisedT{~EDIK~}}&\revisedT{Ours} &       &\revisedT{44.1}& \\
          &  \revisedT{RPL}&  & \revisedT{43.5}&\revisedT{0.6} \\
          &  \revisedT{FPL}&  & \revisedT{43.2}&\revisedT{0.9} \\
          &  \revisedT{SPL}&  & \revisedT{42.7}&\revisedT{1.4} \\
          \hline
           \multirow{4}[0]{*}{\revisedT{~LDSK~}}& &  \revisedT{Ours}     & \revisedT{48.1}& \\
          &  &   \revisedT{w/o EMA}    & \revisedT{47.1}&\revisedT{1.0} \\
          &  &  \revisedT{SP}    & \revisedT{44.5}&\revisedT{3.6} \\
          &  &   \revisedT{MUP}    & \revisedT{43.4}&\revisedT{4.7} \\
          \bottomrule
    \end{tabular}%
    }
  \label{tab5}%
\end{table}%
\subsection{Ablation Study}
To evaluate the effectiveness of each component in the proposed framework, we conduct a comprehensive set of ablation experiments \revised{in the GTA5 $\rightarrow$ Cityscapes adaptation scenario}, as presented in Table~\ref{tab3} and Table~\ref{tab5}. Precisely, we conduct our analysis in terms of EDIK and LDSK.

\begin{figure}[h]
	\centering
	\includegraphics[width=3.5in]{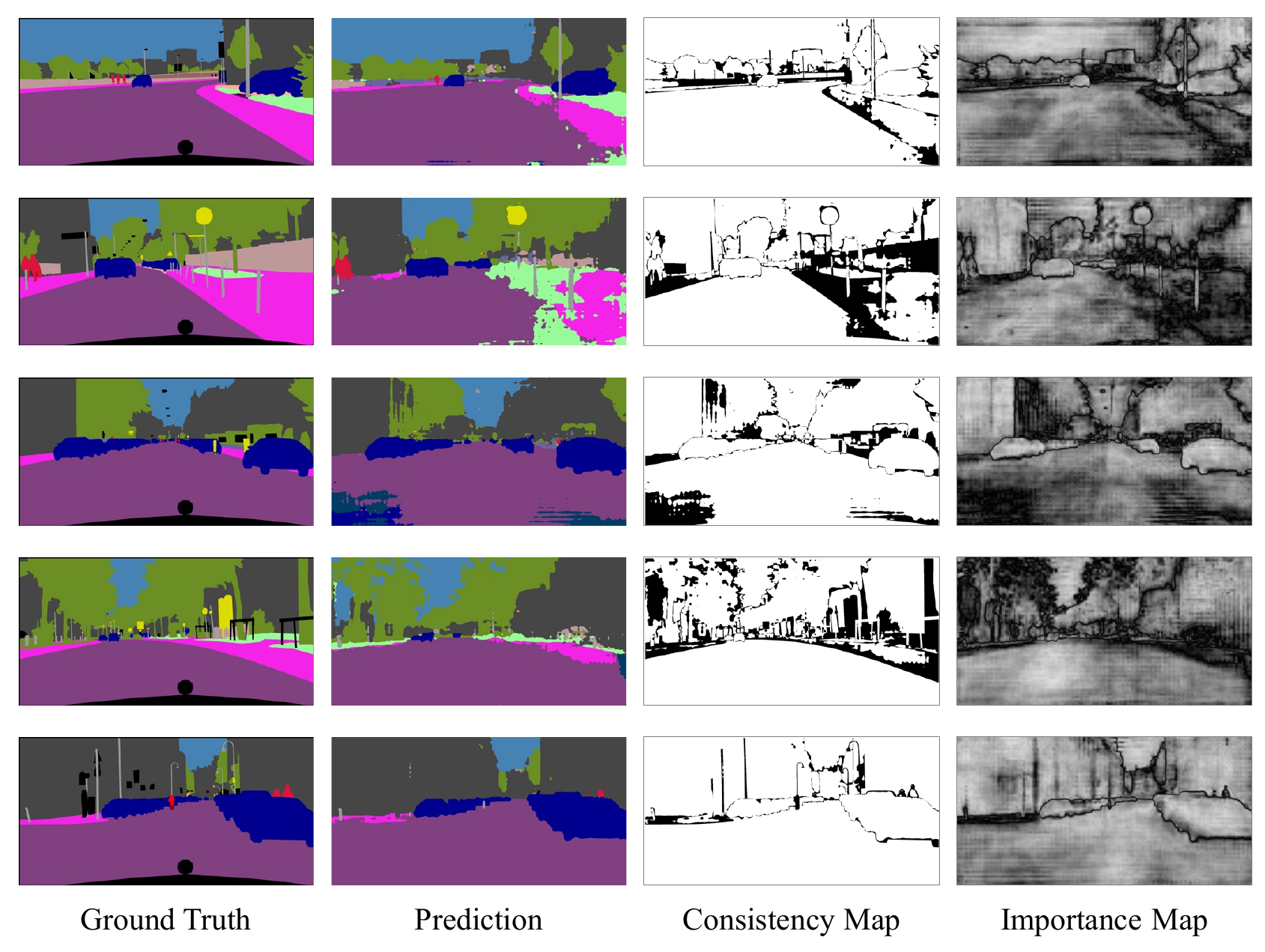}
	\caption{Visualization results of representative examples \revised{on GTA5 $\rightarrow$ Cityscapes} using the proposed importance calculation strategy. The prediction in the second column is inferred from the well-trained source model (best viewed in color).}
	\label{fig5}
\end{figure}

\revised{
\subsubsection{EDIK}
we investigate the importance-aware-based pseudo-label loss $\mathcal{L}_{IA}$ and information maximization loss $\mathcal{L}_{IM}$ (IM) as shown in Tabel~\ref{tab3}.} $\mathcal{L}_{IA}$ extracts the domain-invariant knowledge between the source and target domains to improve the classification ability, improving by $31.2\%$ in mIoU over the ``Source only''. $\mathcal{L}_{IM}$ effectively improves the adaptation ability of the well-trained source model in the target domain by encouraging peak probability distribution of predictions. We visualize the consistency map and the importance map of some representative examples after passing through the well-trained source model in Fig.~\ref{fig5}.
The consistency map is obtained by extracting the consistent regions of the ground truth and the prediction.
These examples demonstrate the consistency of the importance distribution map and prediction mask. This indicates that our designed method for estimating the importance of each prediction effectively guides the model to mitigate the influence of domain shift during training.
\revised{We further investigate the impact of the importance-aware mechanism and the results are shown in the left part of Table~\ref{tab5}. ``RPL" indicates that we replaced the $\mathcal{L}_{IA}$ with a regular pseudo-label loss. ``FPL" and ``SPL" indicate that the substitution of importance $\omega$ in $\mathcal{L}_{IA}$ with the first largest probability $(p^1)$ and second largest probability $(1-p^2)$ respectively. %
\revisedT{Compared with them, the proposed importance-aware mechanism effectively reduces the domain shift. This is achieved by suppressing biased noise predictions under the guidance of the importance distribution map.}
It demonstrates the IA mechanism serves as an effective solution for addressing the lack of labels in the target domain in SFDA.
}

\begin{figure}[!t]
	\centering
	\includegraphics[width=3.5in]{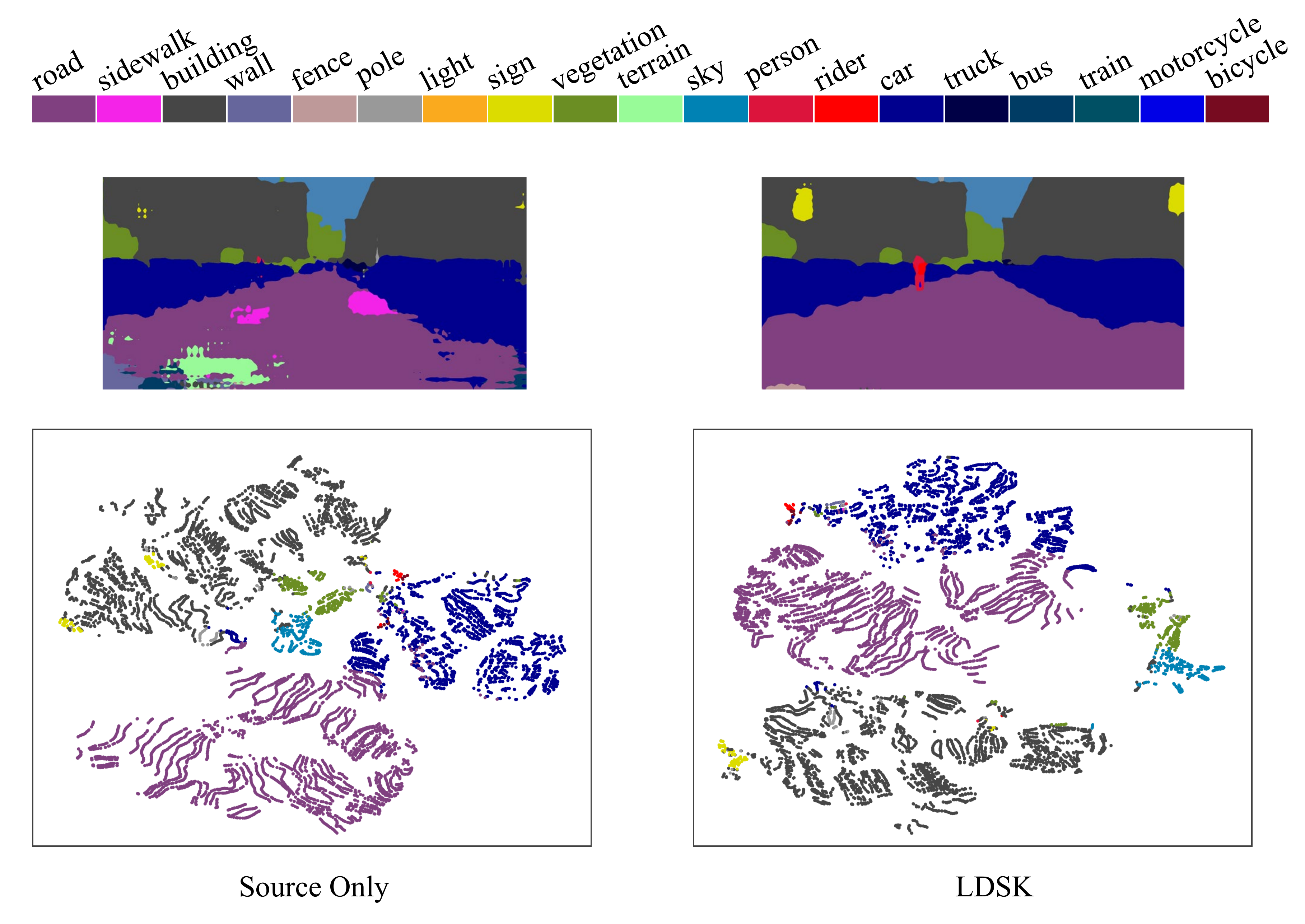}
	\caption{When adapting from GTA5, visualize the feature distribution of a single image from Cityscapes by assigning labeled classes with corresponding colors and using t-SNE to plot the distribution (best viewed in color).}
	\label{fig6}
\end{figure}

\revised{
\subsubsection{LDSK} As shown in Tabel~\ref{tab3}, we evaluate the effectiveness of the prototype-symmetric cross-entropy loss $\mathcal{L}_{PS}$ and the prototype-enhanced cross-entropy loss $\mathcal{L}_{PE}$.}
They improve the self-learning ability within the target domain by fully utilizing the information of the target prototype.
Compared with the source model, $\mathcal{L}_{PS}$ and $\mathcal{L}_{PE}$ show an improvement of $15.3\%$ and $30.3\%$, respectively.
\revisedT{Especially regarding the improvement of $\mathcal{L}_{PE}$, it demonstrates the effectiveness of our proposed PC strategy. This strategy is better suited for completing the adaptation compared to existing methods that use the source model for filtering.}
Furthermore, we use t-SNE~\cite{VanDerMaaten_2008_VisualizingData} to obtain the feature distribution in low dimensions as shown in Fig.~\ref{fig6}.
LDSK can successfully guide features to form tight clusters, thereby improving the discriminativeness of the features. 
Particularly, the features of the ``sign'' class (represented in yellow) exhibit a tighter clustering under the guidance of the prototype-contrast strategy. As a result, accurate segmentation of this class is achieved in the actual prediction map.
\revised{In LDSK, the memorized model using EMA and the dynamic prototypes derived from a single image are designed to facilitate the acquisition of domain-specific knowledge. In the right part of Table~\ref{tab5}, we designed additional experiments to investigate their effectiveness. ``SP'' denotes the utilization of static prototypes, while ``MUP" indicates the adoption of a momentum update approach for prototype establishment following~\cite{Qiu_2021_SourcefreeDomain, Hegde_2021_AttentivePrototypes}. In the ``MUP'' scenario, prototype updates are governed by ${k^c_m}^{(i)} = 0.99{k^c_m}^{(i-1)} + 0.01{k^c_m}^{(i)}$ at every iteration $i$. Directly using the target model to construct prototypes resulted in a $0.6$ decrease in mIoU due to the noise of the current data. We observed a significant performance drop for ``SP'' and ``MUP''. The reason behind this decline is that the established prototypes of our IAPC framework primarily serve to learn target-specific information through the local contrast learning strategy. Dynamic prototypes derived from individual images are more advantageous for the model in extracting detailed information from the current target data.
}

These components complement each other, and their combination achieves the best performance.

\vspace{-2pt}
\subsection{Hyperparameter Analysis}
Our proposed IAPC framework utilizes four hyperparameters as weights to control the strength of four losses $\mathcal{L}_{IA},\mathcal{L}_{PE},\mathcal{L}_{PS},\mathcal{L}_{IM}$.
In this part, we conduct a sensitivity analysis of these weights \revised{in the GTA5 $\rightarrow$ Cityscapes adaptation scenario}, as presented in Table~\ref{tab4}.
Each row in the table corresponds to a separate experiment in which we varied the value of a specific hyperparameter while keeping the others at their optimal values. 
Different weights of PE lead to some fluctuations in the segmentation performance.
As the weight value increases, we observe a significant performance drop, which is attributed to the detrimental error growth caused by excessively large PE weight. Apart from the IM, the proposed IPAC solution is generally not sensitive to the other parameters, and the performance remains relatively stable across different weight values.

\begin{table}[htbp]

\arrayrulecolor{black}
\renewcommand{\arraystretch}{1.1}
  \centering
  \caption{Hyperparameter sensitivity analysis \revised{of the GTA5 $\rightarrow$ Cityscapes adaptation scenario}. The best values and results are shown in bold.}
    \vspace{-5pt}
    \setlength{\tabcolsep}{3.5mm}{
    \begin{tabular}{cccc|c}
    \toprule
    \revised{$\mathcal{L}_{IA}$}&\revised{$\mathcal{L}_{PE}$}&\revised{$\mathcal{L}_{PS}$}&\revised{$\mathcal{L}_{IM}$}&\revised{mIoU}\\
     \hline
    \revised{0.1}&\multirow{4}{*}{\revised{0.5}}&\multirow{4}{*}{\revised{0.04}}&\multirow{4}{*}{\revised{2.0}}&\revised{49.1}\\
    \revised{\textbf{0.2}}&&&&\revised{\textbf{49.4}}\\
    \revised{0.3}&&&&\revised{48.8}\\
    \revised{0.4}&&&&\revised{48.8}\\
    \hline
    \multirow{5}{*}{\revised{0.2}}&\revised{0.3}&\multirow{5}{*}{\revised{0.04}}&\multirow{5}{*}{\revised{2.0}}&\revised{49.2}\\
    &\revised{0.4}&&&\revised{49.3}\\
    &\revised{\textbf{0.5}}&&&\revised{\textbf{49.4}}\\
    &\revised{0.6}&&&\revised{49.0}\\
    &\revised{0.7}&&&\revised{47.6}\\
    \hline
    \multirow{5}{*}{\revised{0.2}}&\multirow{5}{*}{\revised{0.3}}&\revised{0.02}&\multirow{5}{*}{\revised{2.0}}&\revised{48.9}\\
    &&\revised{0.03}&&\revised{48.9}\\
    &&\revised{\textbf{0.04}}&&\revised{\textbf{49.4}}\\
    &&\revised{0.05}&&\revised{49.3}\\
    &&\revised{0.06}&&\revised{49.0}\\
    \hline
    \multirow{5}{*}{\revised{0.2}}&\multirow{5}{*}{\revised{0.3}}&\multirow{5}{*}{\revised{0.02}}&\revised{1.0}&\revised{48.5}\\
    &&&\revised{1.5}&\revised{48.9}\\
    &&&\revised{\textbf{2.0}}&\revised{\textbf{49.4}}\\
    &&&\revised{2.5}&\revised{49.1}\\
    &&&\revised{3.0}&\revised{49.2}\\
    \bottomrule
    \end{tabular}%
    }
  \label{tab4}%
\end{table}%

\vspace{-8pt}
\subsection{Failure Case Analysis}
\begin{figure}[t!]
    \centering
    \includegraphics[width=3.5in]{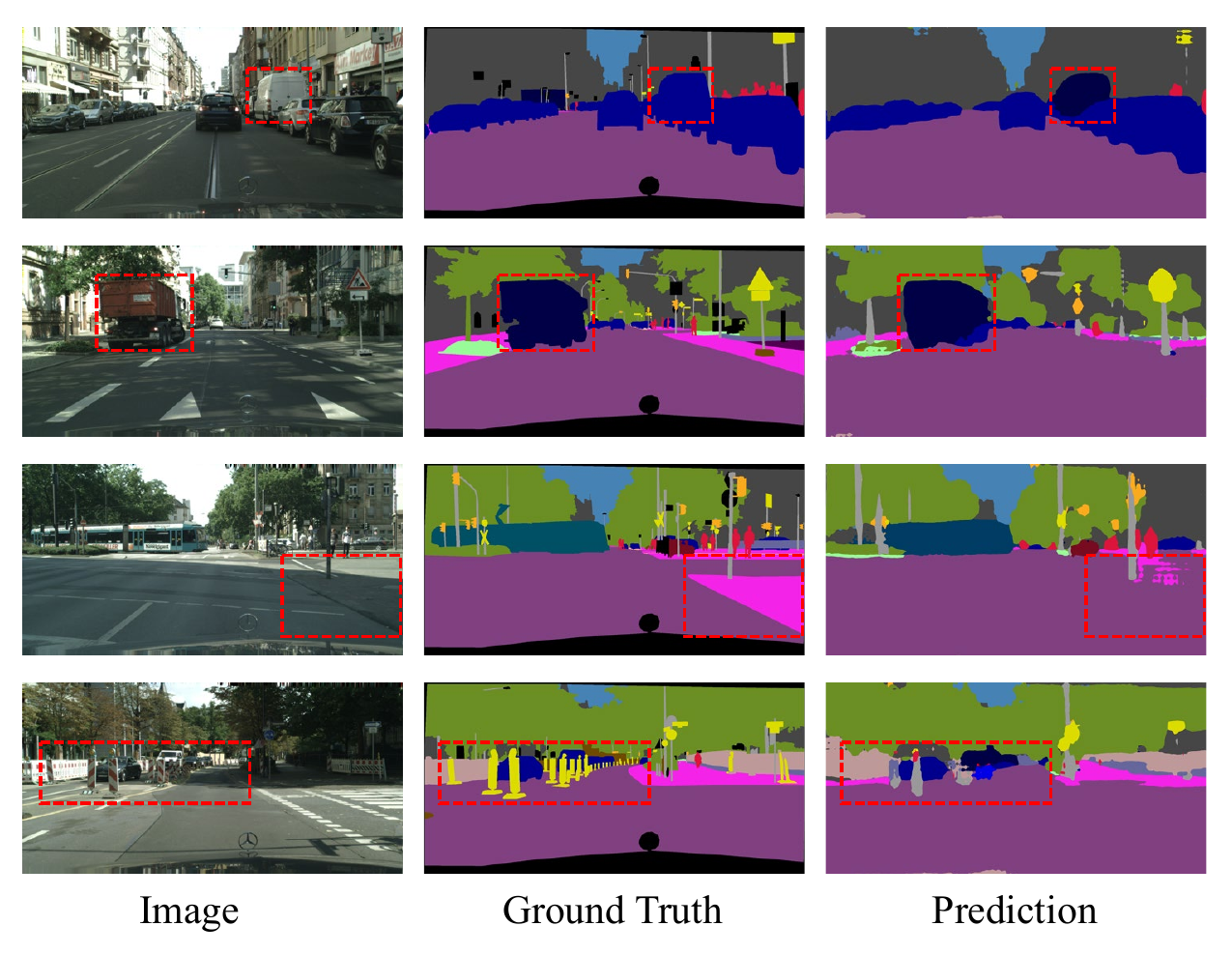}
    \caption{Examples of prediction failures \revised{on GTA5 $\rightarrow$ Cityscapes}. They are commonly caused by challenges in class discrimination, variations in lighting, and the presence of less-frequent objects (best viewed in color and with zoom).}
    \label{fig7}
\end{figure}

For a more comprehensive analysis of the performance of our IPAC, we showcase several examples of prediction failures \revised{on GTA5 $\rightarrow$ Cityscapes, as shown} in Fig.~\ref{fig7}. 
In the first row, a larger-sized car is incorrectly identified as a truck due to the similarity of its square model to a truck. 
In the second row, although the truck is correctly classified, the front bottom of the truck is affected by the low-light environment and the steering of the wheels, causing the model to mistakenly assume that this part is the car in front of the truck. In the third row, we observe that the continuous shadows cover most of the road and the sidewalk, and the road in the middle-right section is not clearly visible.
These lighting variations and road conditions that are difficult to distinguish for humans can confuse the model and result in incorrect predictions. In the last row, some infrequent signs fail to predict, whereas the common signs on the right side are correctly segmented.
It is worth noting that these less-frequent object categories often pose challenges for cross-domain semantic segmentation, and this problem becomes more prominent when the source domain is inaccessible.

\vspace{-15pt}
\revisedT{\subsection{Object Detection}}
\revisedT{
To further verify the generalizability of the proposed IAPC, we have applied our method to the SFDA object detection task.
Following~\cite{li2021free_SFOD, vibashan2023instance_IRG}, we adopt Faster-RCNN with ResNet50 as the detection model. The training strategy of IAPC was referenced from~\cite{vibashan2023instance_IRG}, with the exception that the weights of the three losses $\mathcal{L}_{IA},\mathcal{L}_{PS},\mathcal{L}_{PE}$ were set to $1$, $0.4$, and $1$, respectively. 
We evaluate the proposed IAPC over the commonly used Cityscapes $\rightarrow$ FoggyCityscapes object detection scenario, as presented in Table~\ref{tab_ob}. 
The mean Average Precision (mAP) with an IoU threshold of $0.5$ was reported. 
From Table~\ref{tab_ob}, it can be observed that the proposed IAPC is also effective for the object detection task. 
While our method does not achieve the best performance for each category individually, its competitive performance across categories results in an overall excellent performance of $37.6$ in mAP. 
In scenarios where source domain data is not accessible, IAPC surpasses traditional UDA methods~\cite{Chen_2018_CVPRDAFster, Cai_2019_CVPR_MTOR, Speng_2020_CDN,zhuang2020_iFAN, Hsu_2020_WACV_Pro_DA}. 
This demonstrates that our approach sufficiently facilitates domain adaptation while considering data privacy. 
Furthermore, compared to the SFDA method HCL~\cite{Huang_2021_ModelAdaptation}, which is also applicable to the object detection task, IAPC provides an improvement of $3.0$ in mAP. 
Additionally, IAPC surpasses IRG~\cite{vibashan2023instance_IRG}, which leveraged contrastive learning, by $0.5$ in mAP. 
Overall, these results confirm the effectiveness of the IA mechanism and PC strategy designed by this work in facilitating domain-invariant knowledge learning and domain-specific knowledge extraction.}

\begin{table}[!t]
  \centering
  \caption{\revisedT{Experimental results of the Cityscapes $\rightarrow$ FoggyCityscapes object detection adaptation scene in terms of mAP. The \textbf{SF} column indicates whether the adaptive method is source-free. The best results are shown in bold.}}
  \resizebox{1\linewidth}{!}{
    \begin{tabular}{c|p{2.5mm}<{\centering}|p{3.5mm}<{\centering}p{3.5mm}<{\centering}p{3.5mm}<{\centering}p{3.5mm}<{\centering}   p{3.5mm}<{\centering}p{3.5mm}<{\centering}p{3.5mm}<{\centering}p{3.5mm}<{\centering}|c}
    \toprule
    \multicolumn{1}{c}{\revisedT{Methods}} &\multicolumn{1}{c}{\revisedT{\textbf{SF}}}& \rotatebox{45}{\revisedT{Person}} & \rotatebox{45}{\revisedT{Rider}} & \rotatebox{45}{\revisedT{Car}}   & \rotatebox{45}{\revisedT{Truck}} & \rotatebox{45}{\revisedT{Bus}}   & \rotatebox{45}{\revisedT{Train}} & \rotatebox{45}{\revisedT{Motorbike}} & \multicolumn{1}{c}{\rotatebox{45}{\revisedT{Bicycle}}} & \revisedT{mAP} \\
    \midrule
         \revisedT{Source Only} &   \multirow{6}{*}{\revisedT{\XSolidBrush}}     &\revisedT{31.1}  & \revisedT{38.7}  & \revisedT{36.1}  & \revisedT{19.8}  & \revisedT{23.5}  & \revisedT{9.1}   & \revisedT{21.8}  & \revisedT{30.5}  & \revisedT{26.3} \\
    \revisedT{DA Faster \cite{Chen_2018_CVPRDAFster}} &       & \revisedT{25.0}  & \revisedT{31.0}  & \revisedT{40.5}  & \revisedT{22.1}  & \revisedT{35.3}  & \revisedT{20.2}  & \revisedT{20.0}  & \revisedT{27.1}  & \revisedT{27.6} \\
    \revisedT{MTOR \cite{Cai_2019_CVPR_MTOR}}  &       & \revisedT{30.6}  & \revisedT{41.4}  & \revisedT{44.0}  & \revisedT{21.9}  & \revisedT{38.6}  & \revisedT{40.6}  & \revisedT{28.3}  & \revisedT{35.6}  & \revisedT{35.1} \\
    \revisedT{CDN \cite{Speng_2020_CDN}}   &       & \revisedT{35.8}  & \revisedT{45.7}  & \revisedT{50.9}  & \revisedT{30.1}  & \revisedT{42.5}  & \revisedT{29.8}  & \revisedT{30.8}  & \revisedT{36.5}  & \revisedT{36.6} \\
    \revisedT{iFAN DA \cite{zhuang2020_iFAN}} &       & \revisedT{32.6}  & \revisedT{\textbf{48.5}} & \revisedT{22.8}  & \revisedT{\textbf{40.0}} & \revisedT{33.0}  & \revisedT{\textbf{45.5}} & \revisedT{31.7} & \revisedT{27.9}  & \revisedT{35.3} \\
    \revisedT{Progressive DA \cite{Hsu_2020_WACV_Pro_DA}}&       & \revisedT{36.0}  & \revisedT{45.5}  & \revisedT{\textbf{54.4}} & \revisedT{24.3}  & \revisedT{\textbf{44.1}} & \revisedT{25.8}  & \revisedT{29.1}  & \revisedT{35.9}  & \revisedT{36.9} \\
    \midrule
    \revisedT{SFOD \cite{li2021free_SFOD}}  &   \multirow{6}{*}{\revisedT{\Checkmark}}    & \revisedT{21.7}  & \revisedT{44.0}  & \revisedT{40.4}  & \revisedT{32.2}  & \revisedT{11.8}  & \revisedT{25.3}  & \revisedT{34.5}  & \revisedT{34.3}  & \revisedT{30.6} \\
    \revisedT{HCL \cite{Huang_2021_ModelAdaptation}}  &       & \revisedT{26.9}  & \revisedT{46.0}  & \revisedT{41.3}  & \revisedT{33.0}  & \revisedT{25.0}  & \revisedT{28.1}  & \revisedT{\textbf{35.9}} & \revisedT{40.7}  & \revisedT{34.6} \\
    \revisedT{LODS~\cite{li2022source_LODS}}  &       & \revisedT{34.0}  & \revisedT{45.7}  & \revisedT{48.8}  & \revisedT{27.3}  & \revisedT{39.7}  & \revisedT{19.6}  & \revisedT{33.2}  & \revisedT{37.8}  & \revisedT{35.8} \\
    \revisedT{Mean-Teacher \cite{NIPS2017_68053af2_MeanTeacher}}&       & \revisedT{33.9}  & \revisedT{43.0}  & \revisedT{45.0}  & \revisedT{29.2}  & \revisedT{37.2}  & \revisedT{25.1}  & \revisedT{25.6}  & \revisedT{38.2}  & \revisedT{34.3} \\
    \revisedT{IRG \cite{vibashan2023instance_IRG}}  &       & \revisedT{\textbf{37.4}} & \revisedT{45.2}  & \revisedT{51.9}  & \revisedT{24.4}  & \revisedT{39.6}  & \revisedT{25.2}  & \revisedT{31.5}  & \revisedT{\textbf{41.6}} & \revisedT{37.1} \\
    \revisedT{IAPC (Ours)}  & & \revisedT{36.9}  & \revisedT{42.6}  & \revisedT{50.9}  & \revisedT{28.4}  & \revisedT{40.1}  & \revisedT{32.7}  & \revisedT{29.9}  & \revisedT{39.4}  & \revisedT{\textbf{37.6}} \\
    \bottomrule
    \end{tabular}%
    }
  \label{tab_ob}%
\end{table}%

\vspace{12pt}
\section{Conclusion}
In this work, we propose a Source-Free Domain Adaptation (SFDA) method for road-driving scene semantic segmentation, called IAPC.
The proposed Importance-Aware Prototype-Contrast (IAPC) framework enables end-to-end adaptation from a source model to a target domain without accessing the source data and target labels.
We introduce an Importance-Aware (IA) mechanism to extract domain-invariant knowledge from the source model.
Additionally, we introduce a Prototype-Contrast (PC) strategy to learn domain-specific knowledge from unlabeled target domain data.
As a result, we obtain a model with excellent segmentation performance on the target data. 
\revisedT{Experimental results on two benchmarks for domain adaptive semantic segmentation show that our method achieves state-of-the-art performance. Ablation studies verify the effectiveness of each component in IAPC.}

In the future, we intend to explore source-free domain adaptation for panoramic semantic segmentation to enable 360{\textdegree} surrounding perception in the context of intelligent vehicles. %
We will also investigate the potential of our method in other vision tasks such as panoptic segmentation, 3D scene understanding, \textit{etc.}

\bibliographystyle{IEEEtran}
\bibliography{ref.bib}

\end{document}